\documentclass[final,times,5p,twocolumn]{elsarticle}
\usepackage{lineno}
\usepackage{amssymb}
\usepackage{amsmath}
\usepackage[ruled,linesnumbered]{algorithm2e}
\usepackage{xcolor}
\usepackage{mdframed}
\usepackage{amsmath,amssymb,amsfonts}
\usepackage{graphicx}
\usepackage{epstopdf}
\usepackage{subfig}
\usepackage{gensymb}
\usepackage{subfig}
\usepackage{multirow,booktabs,color,soul,threeparttable}
\usepackage{xcolor}
\usepackage{mdframed}
\usepackage{cleveref}
\usepackage{rotating}
\newcommand{\RNum}[1]{\uppercase\expandafter{\romannumeral #1\relax}}
\definecolor{hl}{rgb}{0.75,0.75,0.75}
\sethlcolor{hl}
\definecolor{emph}{rgb}{0,0,1}

\begin{document}

\begin{frontmatter}



\title{Large Language Models as Surrogate Models in Evolutionary Algorithms: A Preliminary Study}


\author[sjtu]{Hao Hao} \ead{haohao@sjtu.edu.cn}
\author[sjtu]{Xiaoqun Zhang} \ead{xqzhang@sjtu.edu.cn}
\author[ecnu,aiedu]{Aimin Zhou\corref{mycorrespondingauthor}} \ead{amzhou@cs.ecnu.edu.cn}\cortext[mycorrespondingauthor]{Corresponding author: Aimin Zhou}

\affiliation[sjtu]{organization={Institute of Natural Sciences},%Department and Organization
            addressline={Shanghai Jiao Tong University}, 
            city={Shanghai},
            postcode={200240}, 
          %   state={Shanghai},
            country={China}}

\affiliation[ecnu]{organization={School of Computer Science and Technology},%Department and Organization
addressline={East China Normal University}, 
city={Shanghai},
postcode={200062}, 
%   state={Shanghai},
country={China}}

\affiliation[aiedu]{organization={Shanghai Institute of AI for Education},%Department and Organization
addressline={East China Normal University}, 
city={Shanghai},
postcode={200062}, 
%   state={Shanghai},
country={China}}

%% Abstract
\begin{abstract}
Large Language Models (LLMs) have achieved significant progress across various fields and have exhibited strong potential in evolutionary computation, such as generating new solutions and automating algorithm design. Surrogate-assisted selection is a core step in evolutionary algorithms to solve expensive optimization problems by reducing the number of real evaluations. Traditionally, this has relied on conventional machine learning methods, leveraging historical evaluated evaluations to predict the performance of new solutions. In this work, we propose a novel surrogate model based purely on LLM inference capabilities, eliminating the need for training.
 Specifically, we formulate model-assisted selection as a classification and regression problem, utilizing LLMs to directly evaluate the quality of new solutions based on historical data. This involves predicting whether a solution is good or bad, or approximating its value. This approach is then integrated into evolutionary algorithms, termed LLM-assisted EA (LAEA). Detailed experiments compared the visualization results of 2D data from 9 mainstream LLMs, as well as their performance on optimization problems. The experimental results demonstrate that LLMs have significant potential as surrogate models in evolutionary computation, achieving performance comparable to traditional surrogate models only using inference. This work offers new insights into the application of LLMs in evolutionary computation. Code is available at: https://github.com/hhyqhh/LAEA.git

\end{abstract}


% \begin{highlights}
% \end{highlights}

%% Keywords

\end{frontmatter}

% \linenumbers

\section{Introduction}

Surrogate-assisted evolutionary algorithms~(SAEAs) serves as a crucial bridge for the practical application of theoretical evolutionary algorithms~(EAs). Surrogate models can effectively reduce the dependency of evolutionary algorithms on the objective evaluation function during the search process, thereby lowering the optimization cost. This is particularly important in expensive optimization problems~(EOPs). The essence of model-assisted evolutionary optimization algorithms lies in approximating the high-cost optimization objective function of a black-box through historical evaluation data, guiding the search process of the evolutionary algorithm.

Surrogate modeling fundamentally serves the purpose of replicating the functionality of complex, resource-intensive evaluations using more cost-effective models~\cite{jin2005comprehensive}. Within this realm, techniques are predominantly split into two categories, namely regression and classification~\cite{hao2020binary}. Normally, regression is leveraged to predict continuous outputs or fitness values correlated with specific inputs. Conversely, classification models are tasked with predicting discrete labels or categories for solutions, an area which is receiving progressively more academic and practical interest. The proliferation of machine learning (ML) technologies~\cite{jordan2015machine} has fueled the adoption of various algorithms in the creation of surrogate models. Notable among these are Gaussian processes (GP)~\cite{liu2013gaussian}, neural networks (NN)~\cite{pan2018classification}, radial basis functions (RBF)~\cite{yuSurrogateassistedHierarchicalParticle2018}, and support vector machines (SVM)~\cite{HaoZZ21}, which have been effectively employed across both single- and multi-objective optimization tasks~\cite{zhou2011multiobjective}. Applications of these techniques span a broad spectrum, from optimizing the design of buildings~\cite{liu2023surrogate} to improving processes in industrial settings such as iron production in blast furnaces~\cite{chugh2017datadrivena}, and the design of neural networks~\cite{9508774}.

The capabilities of large language models (LLMs) across diverse sectors have recently reached impressively advanced stages, as evidenced by various key studies~\cite{min2023recent,lee2023benefits,thirunavukarasu2023large,kasneci2023chatgpt,liu2023summary}. These models operate by assimilating information from extensive text datasets, effectively capturing human knowledge, which enables them to exhibit powerful cognitive functions such as reasoning and decision-making~\cite{Wei0SBIXCLZ22,Wang2023,ZhouSHWS0SCBLC23,YaoZYDSN023}. Given their proficiency in understanding and applying learned knowledge, it is conceivable that LLMs possess insights akin to human experience and pragmatic intuition in the realm of designing optimization algorithms. This prompts a compelling inquiry regarding the utility of LLMs in aiding EAs to confront and solve intricate optimization challenges~\cite{wu2024evolutionary}.

Mainstream SAEAs follow a paradigm wherein historical evaluated solutions are used as training data. Machine learning models are then employed to learn the data distribution and predict the quality of new solutions, thereby guiding the search process of the evolutionary algorithm. This paradigm faces two potential challenges: 
\begin{itemize}
     \item The time and computational costs associated with model training: as the evolutionary algorithm iterates, the model requires frequent updates, leading to repeated training and additional computational overhead.
     \item Limitations related to data types and scales: surrogate models based primarily on Gaussian processes have limited capacity to handle discrete and large-scale data, restricting the application scope of some advanced SAEA methods.
\end{itemize}
Against this backdrop, the emergence of LLMs offers new possibilities for SAEAs. LLMs, a class of deep learning models trained on vast amounts of textual data, possess powerful natural language processing capabilities. They can directly leverage prompts to utilize their inference capabilities to predict the quality of new solutions without the need for separate training. This method can bypass the time and computational costs of model training during the iterative process of the evolutionary algorithm and enhance the model's generalization to data.
 
Motivated by this, our work attempts to use LLMs as surrogate models in SAEAs to directly predict the quality of new solutions. Specifically, the task for LLMs needs to be defined. Based on the literature~\cite{hao2020binary}, model-assisted selection can be viewed as a classification or regression problem. In the classification problem, LLMs need to predict the category of new solutions, such as ``good'' or ``bad''. In the regression problem, LLMs need to predict the value of new solutions. Subsequently, appropriate prompts must be designed to guide LLMs' inference based on the defined task. Finally, LLMs are integrated into SAEAs for the selection of new solutions. We propose LLM-based surrogate models for regression and classification tasks and integrate them into SAEAs, forming the LLM-assisted SAEA (LAEA) algorithm. To validate this approach, we conducted a series of experiments comparing the performance of 9 mainstream LLMs in 2D data visualization and 2 LLMs in optimization problems. The performance was benchmarked against the mainstream SAEA method, Bayesian Optimization (BO), and evaluated in terms of model selection accuracy. Results indicate that LLMs have significant potential in SAEAs, achieving performance comparable to traditional surrogate models solely through prompt engineering. The main contributions of this work are as follows:
\begin{itemize}
     \item An innovative LLM-based surrogate model is proposed for predicting the quality of new solutions in SAEAs, utilizing LLMs' inference capabilities to perform regression and classification tasks without the need for training.
     \item The LLM-assisted SAEA (LAEA) algorithm is introduced, integrating LLM-based surrogate models into SAEAs to facilitate the selection of new solutions.
     \item Detailed experimental analysis is conducted, comparing the performance of LLMs in 2D data visualization and multiple optimization problems, and benchmarking against traditional SAEA algorithms and model selection accuracy.
     \item Extensible open-source code is provided to support further research and development in this area.
\end{itemize}

The remainder of this paper is organized as follows: Section~\ref{sec:preliminaries} presents the preliminaries, including LLM applications in SAEAs, black-box optimization problems, and surrogate model paradigms. Section~\ref{sec:proposed_surrogate} details the implementation of LLMs as surrogate models, including prompt generation for regression and classification tasks and integration into SAEAs. Section~\ref{sec:empircal_stu} describes the experimental design and analysis of results. Section~\ref{sec:conclusion} concludes the paper and discusses future research directions.

\section{Preliminaries}
\label{sec:preliminaries}

LLMs have already found several applications within EAs. This section will provide an overview of these applications, formally define black-box optimization problems, and discuss typical surrogate model paradigms in SAEAs.

\subsection{LLMs Enhanced EAs}
Liu et al.~\cite{wu2024evolutionary} highlight the collaborative benefits of integrating LLMs with EAs. This partnership enhances optimization processes in EAs~\cite{yang2024large,liu2023algorithm,liu2024evolution} through the predictive capacity of LLMs and improves LLM performance by leveraging EA's optimization techniques~\cite{sun2022black,yang2024large,lapid2023open}. This synergy effectively boosts applications ranging from Neural Architecture Search to text generation~\cite{wu2024evolutionary}, illustrating a potent intersection of large language models and evolutionary algorithms. Next, we will list the applications of LLMs in various stages of EAs.
\begin{itemize}
  \item \textbf{Solution generation:} Yang et al.~\cite{yang2023large} pioneered the use of LLMs for evolutionary optimization with their optimization by PROmpting (OPRO) technique, which utilizes LLMs to generate solutions based on natural language descriptions. This approach was extended by Meyerson et al.~\cite{meyerson2023language} through Language Model Crossover (LMX), where LLMs create offspring from text-based parent solutions. Further developments by Liu et al.~\cite{liu2023largeb, liu2023largea} and Bradley et al.~\cite{bradley2023quality} demonstrated LLMs' effectiveness in generating diverse and high-quality solutions in both single and multi-objective optimization tasks, showcasing their potential to enhance evolutionary algorithms by leveraging their generative and pattern recognition capabilities.
  \item \textbf{Algorithm generation:} Wu et al.~\cite{wu2023llm} developed AS-LLM to recommend optimal EAs by analyzing problem and algorithm features using LLMs. In algorithm generation, Pluhacek et al.~\cite{pluhacek2023leveraging} used LLMs to design hybrid swarm intelligence algorithms, while OptiMUS~\cite{ahmaditeshnizi2023optimus} automated all stages of mixed-integer linear programming problems. Liu et al.~\cite{liu2023algorithm, liu2024example} integrated LLMs into the evolutionary process to create new algorithms, demonstrating superior performance in tasks like the traveling salesman problem. Bradley et al.~\cite{bradley2024openelm} introduced OpenELM, a library that facilitates the design of EAs using LLMs for generating variations and evaluating solutions. These developments highlight LLMs' pivotal role in enhancing and innovating EA methodologies.
\end{itemize}
While LLMs have demonstrated remarkable performance in the field of evolutionary algorithms, their application in model-assisted optimization, specifically in the construction of surrogate models, remains largely unexplored. The successes of existing works and the gaps in the model-assisted selection stage have motivated us to investigate the use of LLMs as surrogate models in evolutionary algorithms.

\subsection{Black-box Optimization}

Consider a black-box function $f: \mathbb{R}^n \rightarrow \mathbb{R}$, where each evaluation of $f$ is assumed to be expensive in terms of computational resources or time. The objective of the black-box optimization problem is to find~\cite{DBLP:conf/kdd/GolovinSMKKS17}:

\begin{equation}
\mathbf{x}^* = \arg\min_{\mathbf{x} \in \mathcal{X}} f(\mathbf{x})
\end{equation}
where:
\begin{itemize}
  \item $\mathbf{x} \in \mathbb{R}^n$ represents a vector of decision variables.
  \item $\mathcal{X} \subseteq \mathbb{R}^n$ denotes the feasible region within the decision variable space.
  \item $f(\mathbf{x})$ is the value of the objective function for a given vector $\mathbf{x}$, which is costly to evaluate.
  \item $\mathbf{x}^*$ is the optimal solution, satisfying $f(\mathbf{x}^*) \leq f(\mathbf{x})$ for all $\mathbf{x} \in \mathcal{X}$.
\end{itemize}
The function $f$ is considered a black box as its explicit form is unknown, rendering gradient-based optimization methods inapplicable. Evaluating $f(\mathbf{x})$ typically demands substantial computational effort. The challenge lies in identifying the optimal $\mathbf{x}^*$ with the fewest possible evaluations of $f$, owing to the significant cost associated with each evaluation.

\subsection{Surrogate Model Paradigm}
\label{sec:surrogate_model_paradigm}
Surrogate models serve as approximations for the optimization of black-box functions that are costly to evaluate. A diverse range of supervised machine learning algorithms are typically employed in the construction of surrogate models, including Gaussian processes~\cite{liu2013gaussian}, neural networks~\cite{pan2018classification,hao2021approximated}, radial basis networks~\cite{li2020surrogate}, support vector machines~\cite{hao2020binary}, and others. The basic paradigms of surrogate models have evolved over time, moving from simple replacements of the original black-box function to more sophisticated forms such as approximating objective values, class prediction, and even direct relationship prediction. The specific operational paradigms of surrogate models in EAs can be categorized into three major types:
\begin{itemize}
\item Regression-based: the model $\mathcal{M}_{\text{reg}}$ predicts a continuous output $y$ based on the input vector \( \mathbf{x} \). The general form can be depicted as:
  \begin{equation}
  y = \mathcal{M}_{\text{reg}}(\mathbf{x})
  \label{eq:regression}
  \end{equation}
where \( \mathcal{M}_{\text{reg}} \) represents the regression model, \( \mathbf{x} \) denotes decision variables in the evolutionary algorithm, and \( y \) denotes the predicted output, often serving as a surrogate to expensive black-box functions. In this paradigm, Jones et al.~\cite{jones1998efficient} proposed a Gaussian process-based surrogate model, coupled with expected improvement as the acquisition function, aimed at global optimization within the GA framework. Liu et al.~\cite{liu2014gaussian} employed Gaussian process surrogates in differential evolution algorithms to enhance convergence speed. Similarly, Li et al.~\cite{li2020surrogate} incorporated radial basis functions in particle swarm optimization to improve performance in high-dimensional problems. The use of regression models in multi-objective optimization has resulted in significant performance improvements as demonstrated by Chugh et al.~\cite{chugh2016surrogate} and Song et al.~\cite{song2021kriging}.

\item Classification-based: the model $\mathcal{M}_{\text{cla}}$ aims to map an input vector \( \mathbf{x} \) to a discrete class label \( l \). The general form of the model can be represented as:
    \begin{equation}
      l = \mathcal{M}_{\text{cla}}(\mathbf{x})
      \label{eq:classification}
      \end{equation}
where \( \mathcal{M}_{\text{class}} \) denotes the classification model, \( \mathbf{x} \) represents decision variables in the evolutionary algorithm, and \( l \) is the predicted class label, used to indicate the quality of a solution. Zhou et al.~\cite{zhou2019fuzzy} utilized a fuzzy KNN classifier to filter the quality of solutions, enhancing the performance of the differential evolution~(EA) algorithm. Similarly, Wei et al.~\cite{wei2020classifier} employed a gradient boosting classifier to predict the quality of solutions in a level-based learning swarm optimizer. Classification models also find ample application in multi-objective optimization contexts; Zhang~\cite{zhangPreselectionClassificationCase2018} and Pan~\cite{pan2018classification} respectively applied KNN and neural network classifiers within multi-object evolutionary algorithms to accelerate convergence.
  
\item Relation-based: the model $\mathcal{M}_{\text{rel}}$ focuses on learning the relative advantages directly among solutions. The relation model can be expressed as:
\begin{equation}
  r = \mathcal{M}_{\text{rel}}(\mathbf{x}_1, \mathbf{x}_2)
  \label{eq:relation}
  \end{equation}
  where \( \mathcal{M}_{\text{rel}} \) denotes the relation model, \( \mathbf{x}_1 \) and \( \mathbf{x}_2 \) represent two solution vectors, and \( r \) indicates the predicted relation between the two solutions. As a novel class of surrogate models, fast development has been witnessed this year: Hao et al. \cite{hao2020binary, hao2024enhancing} utilized direct relation learning and prediction to boost the performance of differential evolution and estimation of distribution algorithms in single-objective problems. In the field of multi-objective optimization, the dominance relationships between solutions are directly employed in constructing surrogate models, thus enhancing the performance of algorithms, as reported by Yuan et al.~\cite{yuan2021expensive}, Hao et al.~\cite{hao2022expensive}, and Tian et al.~\cite{hao2022expensive}.
\end{itemize}

Among the three typical paradigms mentioned above, regression and classification are two fundamental methods that will be employed to construct the inference tasks for LLM-based surrogate models. In this paper, we investigate how to utilize LLMs as surrogate models to directly predict the quality of new solutions.

\section{LLMs-assisted Evolutionary Algorithms}
\label{sec:proposed_surrogate}
This section first introduces the basic framework of employing LLMs as surrogates to assist evolutionary algorithms. Subsequently, then details how to use LLMs through prompts for classification and regression tasks, including specific implementation details. Finally, the surrogate will be integrated into an efficient evolutionary algorithm~\cite{hao2024model}.

\subsection{Framework}

\begin{figure*}[htbp]
  \centering
  \includegraphics[width=1\linewidth]{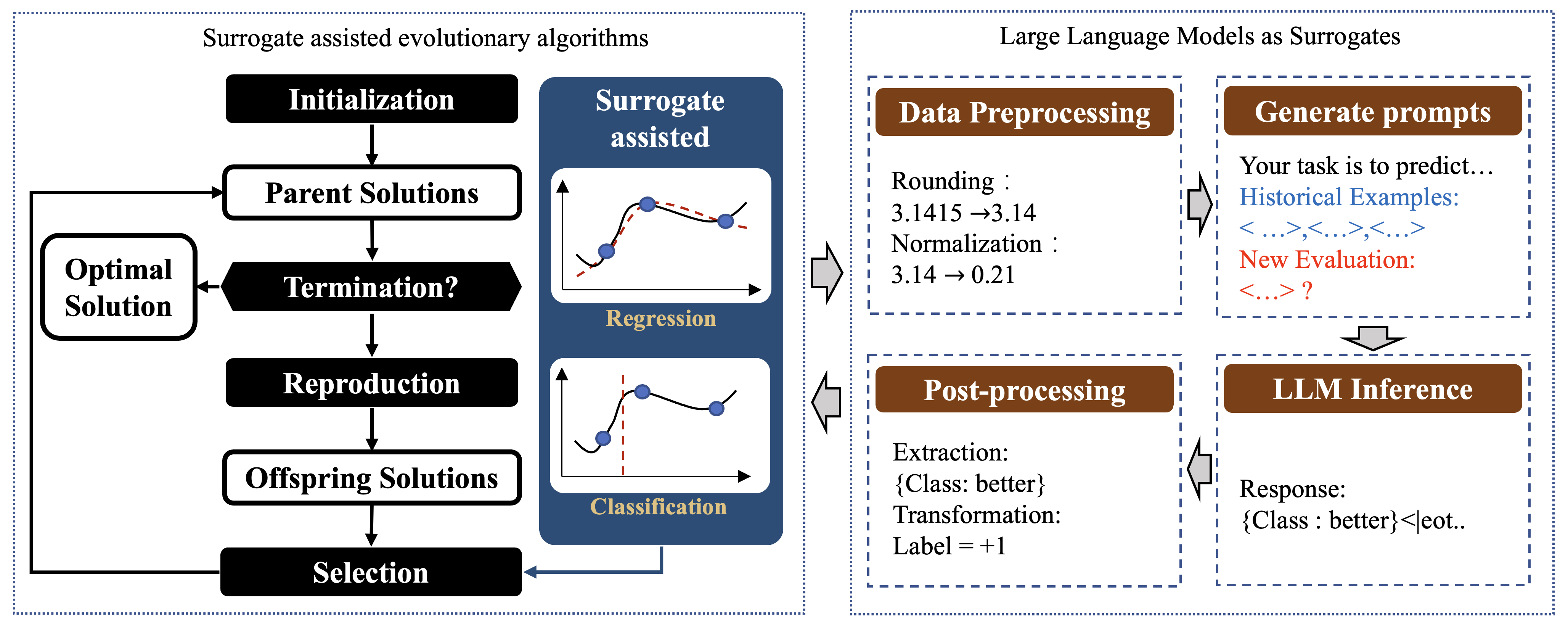}
  \caption{The framework of LLM-assisted evolutionary algorithm~(LAEA). The left side illustrates the basic structure of a surrogate-assisted evolutionary algorithm, where selection can be either a classification or regression task. The right side depicts the four steps of utilizing LLMs as surrogate models: preprocessing, generating prompts, inference, and post-processing.}
  \label{fig:framework}
\end{figure*}

Figure~\ref{fig:framework} presents an overview of the LLM-assisted EA framework. On the left, a typical surrogate-assisted evolutionary algorithm structure is shown. The process begins with the initialization of a population, where a parent population generates offspring through reproduction operators. Subsequently, under model-assisted selection, the next generation of the population is formed. This procedure iterates until termination criteria are met, ultimately outputting the optimal solution. The surrogate-assisted selection task may involve classification or regression tasks, as demonstrated in section~\ref{sec:surrogate_model_paradigm}. On the right, the process of using LLMs as surrogate models is depicted in four steps. Initially, input data undergo preprocessing, which includes rounding and normalization, among other adjustments. Next, prompts describing the task are generated. Then, using the LLM combined with these prompts, inference is conducted to produce the final output. Finally, post-processing is applied where predictive results are extracted from the LLM’s inference and converted into the appropriate format, such as labels for classification tasks or numerical values for regression tasks. The integration of these two processes forms the LLM-assisted EA framework, which is detailed in Algorithm~\ref{alg:llm_ea}.

\subsection{LLMs as Surrogate Models}

The basic process of incorporating LLMs as surrogate models into Evolutionary Algorithms is illustrated in Figure~\ref{fig:framework}. This process comprises four steps: preprocessing, generating prompts, inference, and post-processing, as depicted in Algorithm\ref{alg:llm}. Next, the details will be introduced in conjunction with specific tasks, including regression and classification.

\begin{algorithm}[htbp]
\caption{LLM as Surrogate Model}\label{alg:llm}
\SetKwInOut{Input}{Input}\SetKwInOut{Output}{Output}
\Input{$X$~(evaluated solutions),\\ 
$Y$~(the values or labels of solutions $X$),\\ 
$U$~(to be predicted solutions),\\ 
$LLM$~(large language model),\\
$Opt$~(``Reg" or ``Cla").}
\Output{$\tilde{Y}$~(Predictions for $U$: values or labels).}
\BlankLine
$\tilde{Y} \leftarrow \emptyset$;\\
$X,U \leftarrow \mathrm{Preprocessing}(X,U)$;\\
\For{$u \in U$}{  \label{alg:llm:loop}
$\mathbf{prompt} \leftarrow \mathrm{GeneratePrompt}(X,Y,u,Opt)$;\\
$response \leftarrow \mathrm{Inference}(LLM,\mathbf{prompt})$;\\
$ y \leftarrow \mathrm{PostProcessing}(response, Opt)$;\\

$\tilde{Y} \leftarrow \tilde{Y} \cup y$;
} \label{alg:llm:endloop}
\end{algorithm}

\subsubsection{Regression Task}

For regression tasks, Algorithm~\ref{alg:llm} receives the evaluated solutions $X$ and their value $Y$, as well as the solutions $U$ to be predicted, along with the large language model ($LLM$) used for inference. Initially, the input data are preprocessed by scaling the feature vectors $X$ and $U$ to a range of $[0,1]$, retaining $\beta$ decimal places (by default, $\beta$ is set to 3), as shown in Equation~(\ref{eq:scaling}). 
\begin{equation}
\begin{aligned}
\tilde{z} = \frac{z - \min(Z)}{\max(Z) - \min(Z)}
\label{eq:scaling} 
\end{aligned}
\end{equation}
where $z$ represents the original feature vector, $Z$ denotes the set of feature vectors, and $\tilde{z}$ is the scaled feature vector. Concurrently, the $Y$ values are also scaled to a range of $[0,1]$, retaining $5$ decimal places. This scaling is performed to ensure that the feature vectors are within a consistent range, facilitating the LLM's inference process. Subsequently, prompts describing the task are generated, as illustrated in Figure~\ref{fig:reg_prompt}. These prompts comprise five parts: task description, process description, historical data , newly evaluated feature vectors ($u$), and an emphasis on the output format to ensure the model returns data in JSON format. This facilitates the extraction of prediction results during the post-processing stage. Then, the LLM performs inference based on the prompts, and finally, the predicted results are extracted from the LLM's inference and reverse-scaled to obtain the predicted values. In Algorithm~\ref{alg:llm}, the aforementioned process is iteratively executed (lines~\ref{alg:llm:loop} to \ref{alg:llm:endloop}) until all the unevaluated solutions $U$ have been evaluated by the LLM.

\begin{figure}[htbp]
\centering
\small
\begin{mdframed}[frametitle={\textbf{Regression task prompt}}]
\textcolor{gray}{
Your task is to predict the numerical value of each object based on its attributes. These attributes and their corresponding values are outcomes of a black box function's operation within its decision space. The target value for each object is determined by a specific mapping from these attributes through the black box function. Your objective is to infer the underlying relationships and patterns within the black box function using the provided historical data. This task goes beyond simple statistical analyses, such as calculating means or variances, and requires understanding the complex interactions between the attributes. Please do not attempt to fit the function using code similar to Python; instead, directly learn and infer the numerical values. \\
}
\textcolor{orange}{
\textbf{Procedure:} \\
1. Analyze the historical data to uncover how attributes relate to the numerical values. \\
2. Use these insights to predict the numerical value for new objects based on their attributes. \\
3. Respond using JSON format, e.g. \{`Value': `approximation result'\} \\
}
\textcolor{blue}{
\textbf{Historical Examples:} \\
Features: $\langle 0.338, 0.531,\ldots, 0.363\rangle$ Value: $0.41148$  \\
Features: $\langle 0.207, 0.598,\ldots, 0.285\rangle$ Value: $0.35745$ \\
... \\
Features: $\langle 0.629, 0.029\ldots, 0.279\rangle$ Value: $0.67179$\\
}  
\textcolor{red}{
\textbf{New Evaluation:} \\
Features: $\langle 0.189, 0.917,\ldots, 0.443\rangle $ \\
} 
\textcolor{gray}{
\textbf{Note:} \\Respond in Json with the format \{`Value':`approximation result'\} only.
}
\end{mdframed}
\caption{Regression task prompt}
\label{fig:reg_prompt}
\end{figure}

\subsubsection{Classification Task}

For classification tasks, Algorithm~\ref{alg:llm} receives evaluated solutions $X$ and their corresponding labels $Y$, as well as unevaluated solutions $U$, utilizing a $LLM$ for inference. The label set $Y$ includes categories 1 and 0, provided by upstream tasks. Similar to the regression tasks, preprocessing involves scaling feature vectors $X$ and $U$ to the range $[0,1]$, retaining $\beta$ decimal places (by default, $\beta$ is set to 3). 
The scaling process is shown in Equation~(\ref{eq:scaling}). Subsequently, prompts describing the task are generated, as illustrated in Figure~\ref{fig:cla_prompt}. These prompts comprise five parts: task description, process description, historical data, newly evaluated feature vectors ($u$), and the output format specification. Then, the LLM performs inference based on the prompts. Finally, the predicted results are extracted from the LLM's inference and parsed to obtain the predicted labels. This process is repeated until all the unevaluated solutions $U$ have been assessed by the LLM.

\begin{figure}[htbp]
\centering \small
\begin{mdframed}[frametitle={\textbf{Classification task prompt}}]
\textcolor{gray}{
You are tasked with evaluating each object based on its numerical attributes to determine its category as `better' or `worse'. These attributes derive from a black box function's decision space, with the assessment of the label based on the post-mapping function values. Your role involves discerning the internal variable relationships of the black box function from provided historical data, moving beyond mere statistical analyses like calculating means and variances. \\
}
\textcolor{orange}{
\textbf{Procedure:} \\
1. Identify patterns in how attributes are categorized. \\
2. Apply these patterns to assess new objects, determining whether its category is better or worse. \\
3. Respond using JSON format, e.g. \{`Class': `result'\} \\
}
\textcolor{blue}{
\textbf{Historical Examples:}\\
Features: $\langle 0.555, 0.881,\ldots, 0.491\rangle $, Class: better \\
Features: $\langle 0.593, 0.515,\ldots, 0.456\rangle $, Class: worse \\
... \\
Features: $\langle 0.253, 0.747, \ldots, 0.475\rangle $, Class: better \\
}  
\textcolor{red}{
\textbf{New Evaluation:} \\
$\langle 0.189, 0.917,\ldots, 0.443\rangle $  better or worse?  \\
} 
\textcolor{gray}{
\textbf{Note:}\\ Respond in Json with the format \{`Class': `result'\} only.
}
\end{mdframed}
\caption{Classification task prompt}
\label{fig:cla_prompt}
\end{figure}

The above section introduces the method of employing Zero-Shot Learning to use a Large Language Model as a surrogate model for evaluating solutions within evolutionary algorithms. As models for classification and regression tasks, historical evaluation data are utilized as context, allowing the LLM to predict each unevaluated individual. Next, we will delve into how to integrate the LLM as a surrogate model within evolutionary algorithms to enhance search efficiency.

\subsection{LLM-assisted Evolutionary Algorithms}

When the LLM can be adapted for both regression and classification tasks, integrating the LLM as a surrogate model with EAs becomes straightforward. We adopt the foundation framework from the distribution estimation algorithm proposed in~\cite{hao2024model}, which enhances search efficiency using unevaluated solutions. This algorithm can easily integrate both regression and classification models. Subsequently, the LLMs is embedded as surrogate models within this framework. This algorithm employs a variable-width histogram(VWH)~\cite{zhouEstimationDistributionAlgorithm2015} as a new solution generation operator to enhance global convergence speed. Additionally, it uses promising solutions predicted by the model, without real evaluation (referred to as unevaluated solutions, $\mathcal{P}_u$), in generating new solutions for the next generation. This approach can effectively improve the population distribution~\cite{hao2024enhancing}. The detailed algorithm procedure is illustrated in Algorithm~\ref{alg:llm_ea}. There are several key steps in this algorithm:
\begin{algorithm}[htbp]
  \caption{LLM-assisted Evolutionary Algorithm}\label{alg:llm_ea}
  \SetKwInOut{Input}{Input}\SetKwInOut{Output}{Output}
  \Input{$N$~(population size),\\ $\tau$~(historical data size), \\ $fes_{max}$(maximum function evaluations)\\ $LLM$~(large language model).}
  \Output{$\mathbf{x}^{*}$~(optimal solution).}
  \BlankLine
  $\mathcal{P} \leftarrow \mathrm{Initialize}(N)$\; \label{alg:llmea:popinit} 
  $\mathcal{A} \leftarrow \mathcal{P}$ \; \label{alg:llmea:archinit}
  $\mathcal{P}_u \leftarrow \emptyset $\; \label{alg:llmea:unevalinit}
  $fes \leftarrow 0$\; \label{alg:llmea:fesinit}
  \While{$fes < fes_{max}$}{ \label{alg:llmea:while}
      $\mathcal{Q} \leftarrow \mathrm{Reproduction}(\mathcal{P}\cup \mathcal{P}_{u}, N)$\; \label{alg:llmea:repro}
      $\tilde{V} \leftarrow \mathrm{Prediction}(LLM,\mathcal{A}_{1:\tau},Q,\text{`Reg'})$ \; \label{alg:llmea:reg}
      $\tilde{L} \leftarrow \mathrm{Prediction}(LLM,\mathcal{A}_{1:\tau},Q,\text{`Cla'})$ \; \label{alg:llmea:cla}
      $\mathbf{q} \leftarrow \mathrm{AssistedSelect}(\mathcal{Q}, \tilde{V})$ \; \label{alg:llmea:select_reg}
      $\mathcal{P}_{u} \leftarrow \mathrm{AssistedSelect}(\mathcal{Q}, \tilde{L})$ \; \label{alg:llmea:select_cla}
      $\mathcal{P} \leftarrow \mathrm{Select}(\mathcal{P} \cup \mathrm{Evaluate}(\mathbf{q}), N)$\;  \label{alg:llmea:select}
      $\mathcal{A} \leftarrow \mathcal{A}\cup \mathbf{q}$\;  \label{alg:llmea:arch}
      $fes \leftarrow fes + 1$\; \label{alg:llmea:fes}
  }
  $\mathbf{x}^{*} \leftarrow \mathrm{argmin}(\mathcal{A})$\; \label{alg:llmea:argmin}
\end{algorithm}

\begin{itemize}
\item \textbf{Initialization} (lines \ref{alg:llmea:popinit}-\ref{alg:llmea:unevalinit}): The initialization of the population $\mathcal{P}$ involves sampling $N$ solutions using the Latin Hypercube Sampling (LHS) method~\cite{mckay2000comparisona}. This initial population is evaluated using a black-box function. Additionally, an archive $\mathcal{A}$, identical to the population, is set up, and an unevaluated population $\mathcal{P}_u$ is initiated as an empty set. The function evaluation counter $fes$ is set to $0$.
\item \textbf{Stop condition} (line \ref{alg:llmea:while}): Checks whether the maximum number of evaluations $fes_{max}$ has been reached. If not, the algorithm continues.
\item \textbf{Reproduction} (line \ref{alg:llmea:repro}): Offspring $\mathcal{Q}$ are generated by applying reproduction operators to the combined population $\mathcal{P}\cup \mathcal{P}_{u}$. This process uses a VWH model proposed within EDA/LS~\cite{zhouEstimationDistributionAlgorithm2015} to model evaluated population $\mathcal{P}$ and unevaluated population $\mathcal{P}_u$. This strategy has been proven effective in prior works~\cite{hao2024model,hao2024enhancing}.
\item \textbf{Prediction} (lines \ref{alg:llmea:reg}-\ref{alg:llmea:cla}): LLMs predicts the values $\tilde{V}$ and class $\tilde{L}$ for offspring solutions $\mathcal{Q}$ based on historical data $\mathcal{A}_{1:\tau}$. For classification tasks, labels of historical data are assigned based on the objective values of the $\mathcal{A}_{1:\tau}$: solutions are first sorted by their function values in ascending order, and labels are then assigned as ``+1'' to the top 30\% of solutions and ``0'' to the remaining solutions up to $\tau$. For regression tasks, the actual function values from the historical data are used directly as target values for prediction. Predictions are performed using Algorithm~\ref{alg:llm}.
\item \textbf{Assisted Selection} (lines \ref{alg:llmea:select_reg}-\ref{alg:llmea:select_cla}): Based on $\tilde{V}$, the best solution $\mathbf{q}$ is selected. Then, based on $\tilde{L}$, selections are made for solutions with a label of +1. If more than $N/2$ solutions are chosen, a random selection of $N/2$ solutions is made to form the unevaluated population $\mathcal{P}_u$.
\item \textbf{Selection}: The solution $\mathbf{q}$ is evaluated using a black-box function, combined with the current population $\mathcal{P}$, and the best $N$ solutions are selected based on the real objective function to form the next generation population $\mathcal{P}$.
\item \textbf{Update}: Solution $\mathbf{q}$ is added to the archive $\mathcal{A}$. The evaluation counter $fes$ is incremented.
\item \textbf{Termination}: The optimal solution $\mathbf{x}^{*}$ is determined as the solution with the minimum objective value in the archive $\mathcal{A}$.
\end{itemize}

We introduce the algorithm referred to as the LLM-assisted evolutionary algorithm~(LAEA), which embeds a LLM as a surrogate model for both regression and classification tasks within an EA. The LAEA leverages optimal solutions $\mathbf{q}$ and a subset of unevaluated solutions $\mathcal{P}_u$ for enhancing the efficiency and demonstrating the effectiveness of LLM as a surrogate model. Additionally, a simplified version of this algorithm, solely utilizing LLM for regression without employing it for classification, is termed as LAEA-Reg. Specifically, during the prediction phase in LAEA-Reg, the LLM is only applied to predict the values $\tilde{V}$ of the solutions. In line~\ref{alg:llmea:select_cla} of Algorithm~\ref{alg:llm_ea}, the predicted values $\tilde{V}$ are used to select the top $N/2$ solutions, ordered by $\tilde{V}$ from lowest to highest, to form the unevaluated population $\mathcal{P}_u$.The LAEA-Reg variant potentially reduces the computational overhead of LLM inference by half. However, it does not facilitate the evaluation of LLM's effectiveness as a classification model. Extensive performance assessments will be presented in Section~\ref{sec:empircal_stu}.

\section{Empirical Studies}
\label{sec:empircal_stu}

As a gap-filling study, we will conduct comprehensive experiments to verify the effectiveness of multiple LLMs as surrogate models for EAs. Initially, we will show the performance and visualization results of nine LLMs on a two-dimensional function using visual methods. Subsequently, we will evaluate the performance of LLMs in selecting promising solutions on 5- and 10-dimensional datasets. Following this, we will compare the proposed LAEA algorithm with mainstream black-box optimization algorithms to demonstrate its performance advantages and disadvantages. Finally, a more detailed set of experiments will analyze the model's effectiveness and time complexity in various scenarios. The experimental results will be presented in the following sections.

\subsection{Experimental Setup}

In this section, we provide an overview of the test instances and the LLMs used in our study, outlining the characteristics of the test instances and describing the LLMs employed for the experiments.

\subsubsection{Test Instances}
\label{sec:test_instances}
We used four well-known benchmark test functions for evaluating the performance of the models: Ellipsoid, Rosenbrock, Ackley, and Griewank~\cite{liu2014gaussian}. The details of these functions are as follows:

\begin{itemize}
  \item \textbf{Ellipsoid function:}
     \begin{equation}
      f(\mathbf{x}) = \sum_{i=1}^n i x_i^2
      \end{equation}
     Interval: $-5.12 \leq x_i \leq 5.12$ \\
     Characteristics: The Ellipsoid function is a convex quadratic function. It is unimodal with its global minimum at the origin.
  \item \textbf{Rosenbrock function:}
    \begin{equation}
     f(\mathbf{x}) = \sum_{i=1}^{n-1} [100 (x_{i+1} - x_i^2)^2 + (1 - x_i)^2]
    \end{equation}
     Interval: $-2.048 \leq x_i \leq 2.048$ \\
     Characteristics: it is a non-convex function used to test the performance of optimization algorithms. The global minimum is inside a long, narrow, parabolic shaped flat valley.
  \item \textbf{Ackley function:}
  \begin{align}
    f(\mathbf{x}) & = -20\exp\left(-0.2 \sqrt{\frac{1}{n} \sum_{i=1}^n x_i^2}\right) \notag \\
    & \quad - \exp\left(\frac{1}{n} \sum_{i=1}^n \cos(2\pi x_i)\right) + 20 + e
\end{align}
     Interval: $-32.768 \leq x_i \leq 32.768$ \\
     Characteristics: it features a nearly flat outer region and a large hole at the center. The global minimum is at the origin.
  \item \textbf{Griewank function:}
  \begin{equation}
   f(\mathbf{x}) = 1 + \frac{1}{4000} \sum_{i=1}^n x_i^2 - \prod_{i=1}^n \cos\left(\frac{x_i}{\sqrt{i}}\right)
  \end{equation}
     Interval: $-600 \leq x_i \leq 600$ \\
     Characteristics: it has many widespread local minima, making it difficult for optimization algorithms to converge to the global minimum quickly. The global minimum is at the origin.
\end{itemize}

\subsubsection{LLMs used in the exceptional study}

% \begin{sidewaystable*}[htbp]
\begin{table*}[htbp]
  \renewcommand{\arraystretch}{1.1}
  \renewcommand{\tabcolsep}{5pt}
  \centering
  \caption{Large Language Models Used in the Study}\label{tab:llm_models}
  \begin{tabular}{l|c|c|c|c|c|c}
      \toprule
      Model & Label & Parameters & Size & Quantization & Organization & Open Source\\
      \midrule
      Llama3-8B~\cite{meta2024llama}      & 8b-instruct-q4\_0 & 8B & 4.7 GB & 4-bit & Meta AI & Yes\\
      Llama3-8B*~\cite{meta2024llama}       & Meta-Llama-3-8B-Instruct & 8B & 16 GB & - & Meta AI & Yes\\
      Mixtral-8x7B~\cite{jiang2024mixtral} & 8x7b-instruct-v0.1-q4\_0 & 8$\times$7B & 26 GB & 4-bit & Mistral AI & Yes\\
      Mistral-7B~\cite{jiang2023mistral}     & 7b-instruct-v0.2-q4\_0 & 7B &4.1 GB & 4-bit & Mistral AI & Yes\\
      Phi-2~\cite{javaheripi2023phi} & 2.7b-chat-v2-q4\_0 & 2.7B & 1.6 GB & 4-bit& Microsoft& Yes\\
      Phi-3~\cite{abdin2024phi3} & 3.8b-mini-instruct-4k-q4\_K\_M & 3.8B & 2.3 GB& 4-bit& Microsoft & Yes\\
      Gemma-7B~\cite{gemma_2024} &  7b-instruct-v1.1-q4\_0 & 7B & 5.0 GB & 4-bit &Google  & Yes\\
      GPT-3.5~\cite{openai2022gpt35}      & gpt-3.5-turbo-0125 & - & - & - & OpenAI & No\\
      GPT-4~\cite{openai2024gpt4}        & gpt-4-turbo-2024-04-09 & - & - & - & OpenAI & No\\
      \bottomrule
  \end{tabular}
\end{table*}
% \end{sidewaystable*}

This work employs a variety of large language models, as detailed in table~\ref{tab:llm_models}. The models from Meta AI are based on the Llama3-8B architecture~\cite{meta2024llama}, which has 8 billion parameters. Two versions of this model are utilized: one is the original, unquantized model with a size of 16 GB, and the other is a 4-bit quantized version that is significantly smaller, occupying only 4.7 GB. From Mistral AI, we utilize two models: Mixtral-8x7B~\cite{jiang2024mixtral} and Mistral-7B~\cite{jiang2023mistral}. The former is a sparse mixture of experts~(MOE) model with 56 billion parameters and a size of 26 GB, while the latter has 7 billion parameters and is 4.1 GB in size. Both models employ 4-bit quantization. For models from Microsoft, we chose the Phi-2~\cite{javaheripi2023phi} and Phi-3~\cite{abdin2024phi3} models. Phi-2 has 2.7 billion parameters and a size of 1.6 GB, while Phi-3 has 3.8 billion parameters and occupies 2.3 GB. Both models use 4-bit quantization. Additionally, Google's Gemma-7B model~\cite{gemma_2024} is employed, which has 7 billion parameters and a size of 5.0 GB, utilizing 4-bit quantization. Finally, two models from OpenAI, GPT-3.5~\cite{openai2022gpt35} and GPT-4~\cite{openai2024gpt4}, are included in this study. The specifics regarding their parameters, size, and quantization are not publicly available. These models collectively represent a broad spectrum of the current state of large language models, providing comprehensive evidence for evaluating LLMs as surrogate models.

All 4-bit quantized models were deployed using Ollama~\cite{ollama2024}, while GPT-3.5 and GPT-4 were accessed through OpenAI's API service. The unquantized version of Llama3-8B was deployed using VLLM~\cite{kwon2023efficient} to achieve improved inference speeds.

\subsection{2D Case Study}
\label{sec:2d_case}

In EAs, the role of the surrogate model largely involves selecting a subset of promising solutions from a given set. This section will examine the ability of multiple models to correctly select such subsets. For generality, we define the task as a label-balanced binary classification task. To align with common machine learning terminology, we refer to historical data as training data~($\mathcal{D}_{train}$) and the data used for evaluation as testing data~($\mathcal{D}_{test}$). In reality, the training data only provides context for inference of the LLMs and is not used for model training.

We choose four test problems provided in Section~\ref{sec:test_instances}, with each problem's decision space set to 2-dimensional. Using latin hypercube sampling, 50 points are sampled to form the training data set~($\mathcal{D}_{train}$). The testing data set consists of 400 sampled points ($\mathcal{D}_{test}$), created by forming a grid with 20 evenly spaced points in each dimension. Classification thresholds are determined by the median value of the function values at $\mathcal{D}_{train}$, categorizing $\mathcal{D}_{train}$ and $\mathcal{D}_{test}$ below the threshold as class ``1'' and above the threshold as class ``0''. These points are used to evaluate the regression and classification capabilities of the LLMs. When testing the LLM for regression, the predicted values are sorted in ascending order, with the top 50\% considered as ``1'' samples and the bottom 50\% as ``0'' samples. For testing LLM for classification , the predicted labels from the model are directly used as the final labels.

Nine LLMs listed in Table~\ref{tab:llm_models} will use prompts tailored for classification and regression tasks to infer the values and categories for the 400 test points. Classification accuracy~($acc$), defined by the equation~(\ref{eq:cla_accuracy}), is used to measure each LLM's performance across different functions:

\begin{equation}
  acc = \frac{\sum_{i=1}^{|\mathcal{D}_{test}|} \delta(l_{\text{pre},i}, l_{\text{real},i})}{|\mathcal{D}_{test}|} \label{eq:cla_accuracy}
  \end{equation}
where $l_{pre,i}$ and $l_{real,i}$ are the predicted and real labels, respectively, and $\delta(\cdot)$ is an indicator function that returns ``1'' if two labels are the same and ``0'' otherwise.

\begin{figure*}[htbp]
\centering
\subfloat[Visualization of predicted labels by LLMs as classification models]{%
\includegraphics[width=0.99\textwidth]{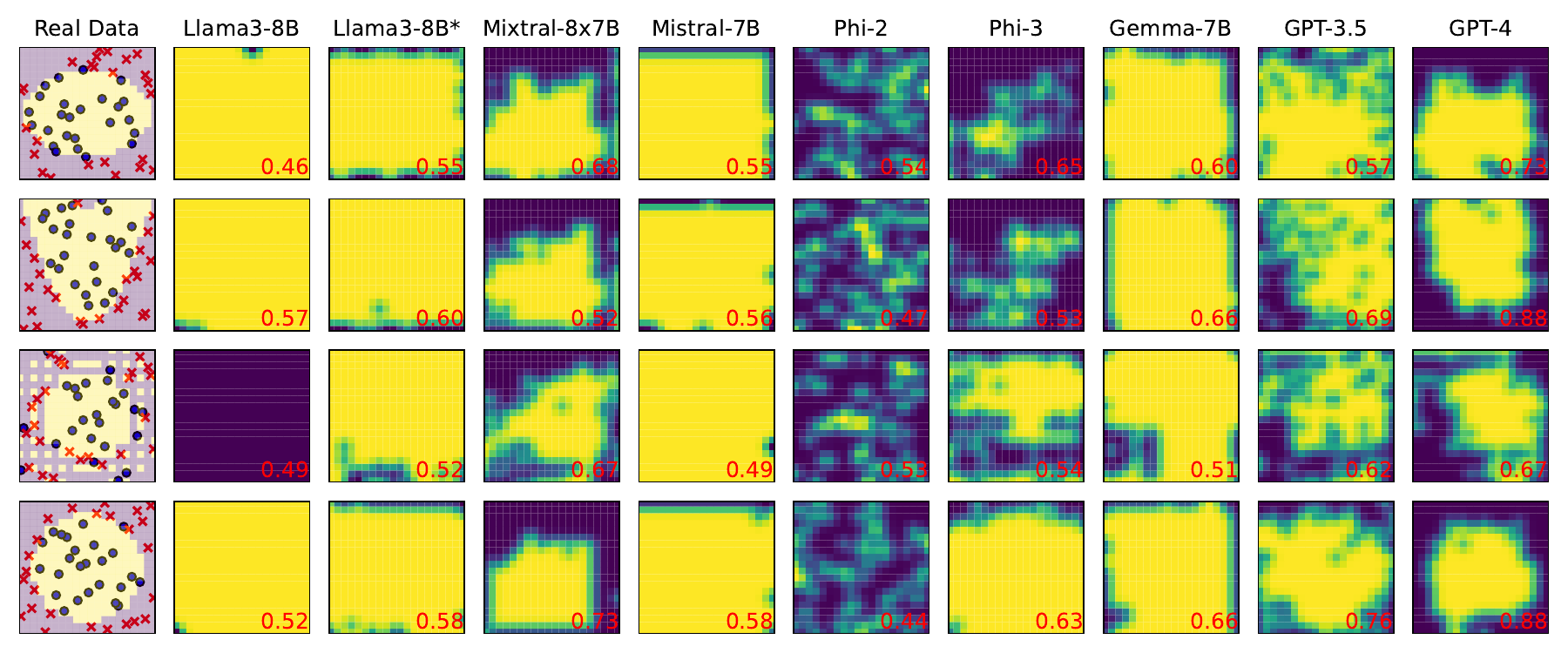}
\label{fig:2d_cla}
} \\
\subfloat[Visualization of predicted labels by LLMs as regression models]{%
\includegraphics[width=0.99\textwidth]{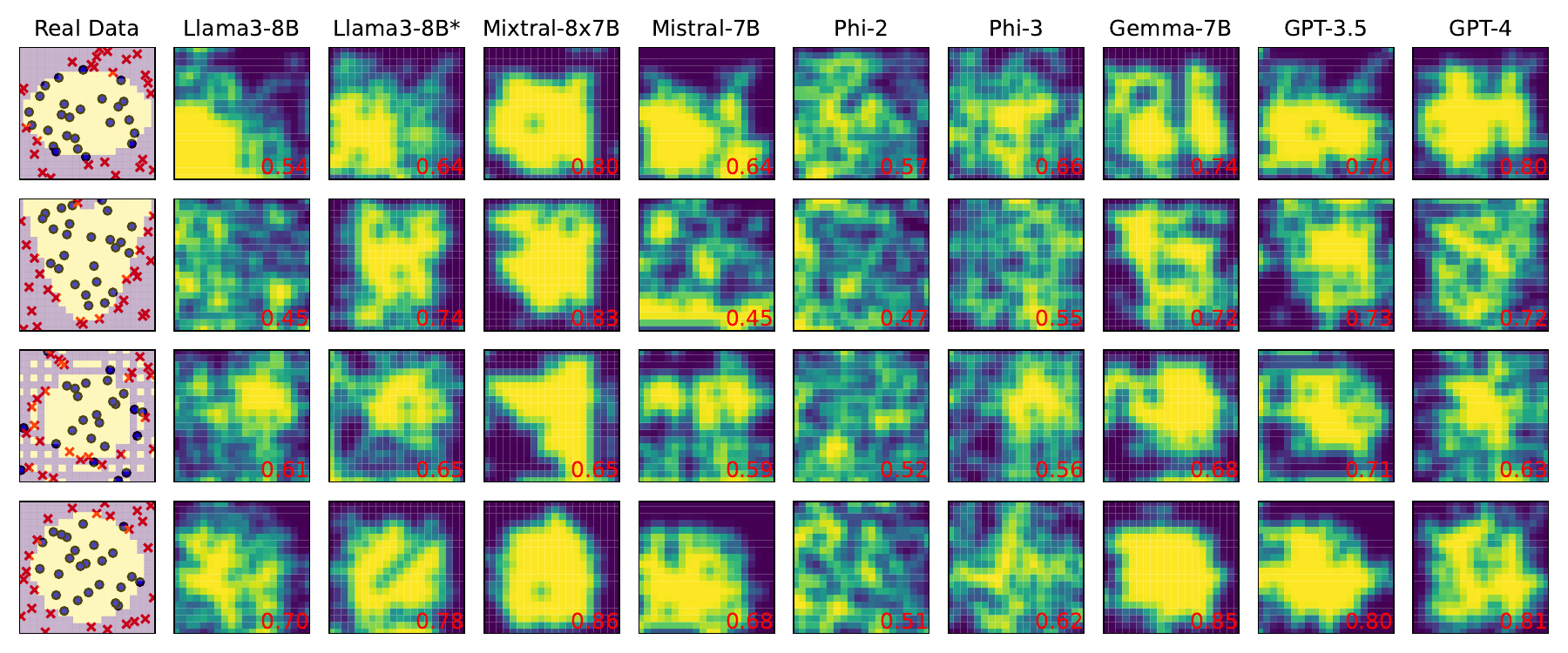}
\label{fig:2d_reg}
}
\caption{Visualization results of label predictions by LLMs for 2-dimensional test problem. The first column represents the real data distribution, where blue $\bullet $ and red $\times$  indicate class ``1'' and class ``0'', respectively. Yellow and purple shading represent the distribution of true labels for the test data. Columns 2 through 10 present the prediction results of nine LLMs, with classification accuracy ($acc$) annotated in the bottom right corner of each test plot.}
\label{fig:2d_plot}
\end{figure*}

\begin{figure}[htbp]
\centering
\includegraphics[width=1\linewidth]{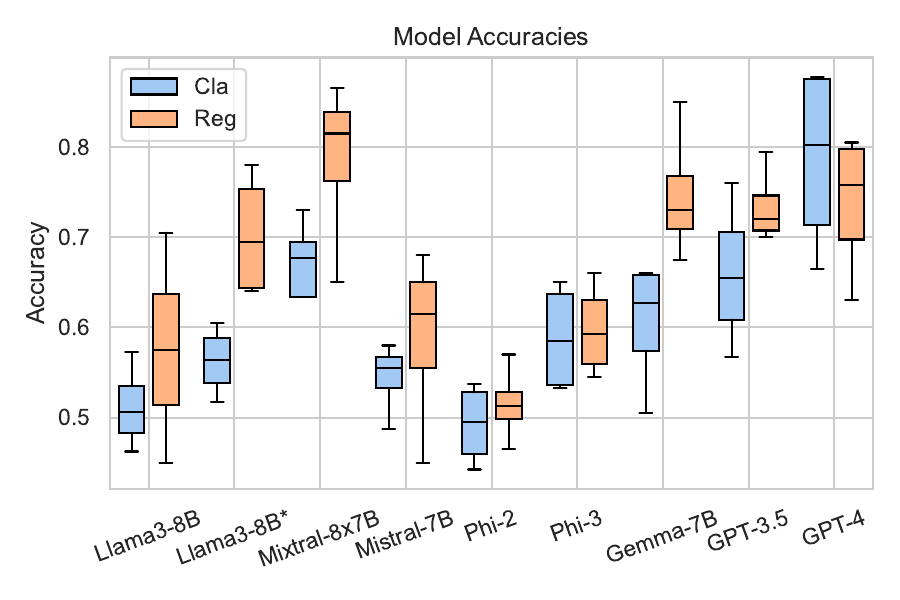}
\caption{Box plots of $acc$ for nine LLMs on four 2D test problems. The blue boxes represent the $acc$ of LLMs used as classification models, while the orange boxes represent the $acc$ of LLMs used as regression models.}
\label{fig:2d_acc}
\end{figure}

The experimental results are illustrated in Figure~\ref{fig:2d_plot}. For LLMs used as classification models (Figure~\ref{fig:2d_cla}), most models can predict labels through contextual inference alone ($acc > 0.5$ ). Notably, the GPT-4, GPT-3.5, Mixtral-8x7B, and Llama3-8B* models capture the distribution of the test data’s original categories and achieve high classification accuracy. For LLMs used as regression models (Figure~\ref{fig:2d_reg}), The Mixtral-8x7B model performs best in the regression tasks, followed by the GPT-4, GPT-3.5 and Llama3-8B* model. These results indicate that some LLMs can serve as surrogate models through inference alone; however, model performance varies across different tasks. Comparing the Llama3-8B and Llama3-8B* models shows that the quantized model exhibits a slight decline in regression prediction capability. Figure~\ref{fig:2d_plot} provides statistical accuracy for various LLMs across the four test problems. The box plot structure and visualization results are generally consistent. Among the open-source models, the Llama3-8B* and Mixtral-8x7B models exhibit the best overall performance, and they perform better on regression tasks compared to classification tasks. By contrast, GPT-4 also demonstrates high accuracy but comes with substantial service costs. Considering these factors, the two open-source models, Llama3-8B* and Mixtral-8x7B, will be the primary models used in subsequent experiments.

\subsection{Selection Acuracy}

Next, we analyze the ability of LLMs to select promising solutions during the iterations of an evolutionary algorithm. The experimental design is as follows: we choose problems with 5 and 10 dimensions and collect population data of parents and offspring generated by the genetic algorithm~(GA)~\cite{sivanandam2008genetic} during the solving process. Specifically, we collect data from the 2nd, 22nd, and 42nd generations to represent the population distribution at the early, middle, and late stages of the algorithm's run. and the process is independently repeated 30 times. This data serves as the training and testing data to overcome the randomness of the evolutionary algorithm population. We test the accuracy of selecting half of the offspring when compared to the accuracy of the true function's selection. The details of using LLMs as regression and classification models for predictions are the same as in the 2D case. We use the precision~(P), recall~(R), and F1 score~(F1) as evaluation metrics, defined as follows:

\begin{itemize}
  \item \textbf{Precision (P)}: The ratio of correctly predicted positive observations to the total predicted positives. It is calculated as:
  \begin{equation}
      P = \frac{TP}{TP + FP}
  \end{equation}
  where $TP$ is the number of true positive observations and $FP$ is the number of false positive observations.
  
  \item \textbf{Recall (R)}: The ratio of correctly predicted positive observations to the all observations in actual class. It is calculated as:
  \begin{equation}
      R = \frac{TP}{TP + FN}
  \end{equation}
  where $FN$ is the number of false negative observations.
  
  \item \textbf{F1 Score (F1)}: The harmonic mean of precision and recall, providing a balance between the two. It is calculated as:
  \begin{equation}
      F1 = 2 \times \frac{P \times R}{P + R}
  \end{equation}
\end{itemize}

where:
\begin{itemize}
  \item $TP$ (True Positives): the number of correctly predicted positive samples, promising solutions that need to be selected.
  \item $FP$ (False Positives): the number of incorrectly predicted positive samples, non-promising solutions that were incorrectly selected.
  \item $FN$ (False Negatives): the number of positive samples that were incorrectly predicted as negative, promising solutions that were incorrectly not selected.
\end{itemize}

\begin{table}[htbp]
\renewcommand{\arraystretch}{1.1}
\renewcommand{\tabcolsep}{2pt}
\caption{Performance metrics of various LLMs across different dimensions. Values are presented as mean (standard deviation).}  \label{tab:selection}
\centering
\begin{tabular}{ccc|ccc}
\toprule
$n$ & LLMs & Task& Precision & Recall & F1-Score \\
\midrule
\multirow{4}{*}{$5$}&Llama3-8B* &Cla & 0.50 (0.07) & 0.48 (0.07) & 0.49 (0.07) \\
&Llama3-8B* &Reg & 0.61 (0.13) & 0.62 (0.13) & 0.62 (0.13) \\
&Mixtral-8x7B &Cla & 0.34 (0.32) & 0.07 (0.09) & 0.11 (0.12) \\
&Mixtral-8x7B &Reg & 0.53 (0.10) & 0.55 (0.12) & 0.54 (0.11) \\
\hline
\multirow{4}{*}{$10$}&Llama3-8B* & Cla & 0.51 (0.08) & 0.49 (0.07) & 0.50 (0.08) \\
&Llama3-8B* & Reg & 0.67 (0.12) & 0.67 (0.12) & 0.67 (0.12) \\
&Mixtral-8x7B & Cla & 0.14 (0.30) & 0.01 (0.03) & 0.02 (0.05) \\
&Mixtral-8x7B & Reg & 0.55 (0.07) & 0.69 (0.21) & 0.60 (0.09) \\
\bottomrule 
\end{tabular}
\end{table}

Llama3-8B* and Mixtral-8x7B models were used for classification~(Cla) and regression~(Reg) tasks, respectively, to predict on 5-dimensional and 10-dimensional datasets. Table~\ref{tab:selection} presents the mean and standard deviation of performance metrics for the two models across four test problems, three stages, and 30 independent runs. The results show that both models perform better on regression tasks compared to classification tasks. An increase in dimension does not significantly affect model performance. The Llama3-8B* model demonstrates stable performance in both regression and classification tasks, whereas the Mixtral-8x7B model performs poorly in classification tasks. Specifically, the recall rate is very low, indicating poor selection capability for solutions.

\begin{figure*}[htbp]
\centering
\includegraphics[width=.8\linewidth]{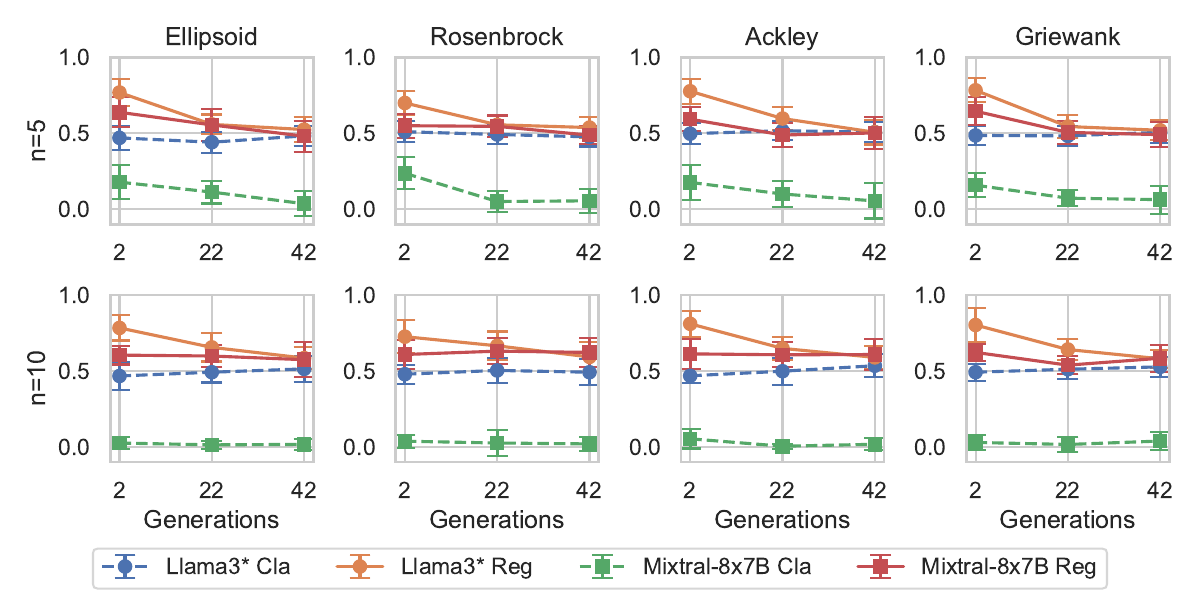}
\caption{Line plots of F1-Score for LLMs during data selection tasks at different stages of the four test problems.}
\label{fig:select_f1_score}
\end{figure*}

To further illustrate the variation in model performance across different test problems and stages, Figure~\ref{fig:select_f1_score} shows the line plots of the F1-Score for LLMs during data selection tasks at different stages of the four test problems. The overall conclusions are consistent with the results presented in Table~\ref{tab:selection}. The Llama3-8B* model demonstrates stable performance across all test problems, while the Mixtral-8x7B model shows poor performance in classification tasks. Additionally, as the iterations of the GA algorithm progress, there is a downward trend in model performance. This decline could be attributed to the convergence properties of the GA, leading to uneven population distribution.

\subsection{Comparative Study}
\label{sec:exp:comparative}

In this section, the Llama3* and Mixtral-8x7B are utilized as surrogate models, embedded within the LAEA algorithm proposed in Algorithm~\ref{alg:llm_ea}. The performance of the proposed algorithm is compared against other mainstream black-box optimization algorithms. The implemented versions of the algorithm in this section include the following four variants:

\begin{itemize} 
  \item LAEA-8B: Utilizes Llama3-8B* as both regression and classification model in the LLM-assisted evolutionary algorithm.
  \item LAEA-Reg-8B: Utilizes Llama3-8B* solely as a regression model in the LLM-assisted evolutionary algorithm.
  \item LAEA-8x7B: Utilizes Mixtral-8x7B as both regression and classification model in the LLM-assisted evolutionary algorithm.
  \item LAEA-Reg-8x7B: Utilizes Mixtral-8x7B solely as a regression model in the LLM-assisted evolutionary algorithm.
\end{itemize}

The comparison algorithms chosen are Bayesian Optimization (BO) and surrogate-assisted evolutionary algorithms, as specified below:
\begin{itemize}
  \item Bayesian Optimization (BO): Incorporates sequential domain reduction~\cite{Standerrobustnesssimpledomain2002a} within the standard BO framework, significantly accelerating the search progress and hastening convergence. It uses a Gaussian process as the surrogate model. For specific implementation details, refer to~\cite{BO2014}.
  \item Surrogate-assisted Evolutionary Algorithm (SAEA): Utilizes various single-objective SAEA provided by the PlatEMO platform~\cite{tian2017platemo}, including SADE~\cite{chen2020surrogate}, SAMSO~\cite{li2020surrogate}, SACOSO~\cite{sun2017surrogate}, and SACC-EAM-II~\cite{blanchard2019surrogate}, representing state-of-the-art surrogate-assisted evolutionary algorithms. All algorithms are executed with default parameter settings.
\end{itemize}
The test functions selected are the four functions from section~\ref{sec:test_instances}, with each test function being evaluated at dimensions of 5 and 10. Each algorithm runs 30 times on each test function, with a maximum of 300 evaluations per run, simulating limited evaluation counts due to high computational cost. The mean optimal values and standard deviations achieved by each algorithm are presented in Table~\ref{tab:comparative}. Additionally, the The Wilcoxon rank-sum test~\cite{hollander2013nonparametric} is used to perform statistical analysis on the optimal values obtained by each algorithm, and the mean rank values achieved by each algorithm are provided.

% \begin{sidewaystable*}[htbp]
\begin{table*}[htbp]
  \renewcommand{\arraystretch}{1.1}
  \renewcommand{\tabcolsep}{1pt}
  \centering
  \caption{Statistics of mean and standard deviation results obtained by nine comparison algorithms on four test functions with $n=5,10$, adhere to a maximum evaluation budget of 300.} \scriptsize
  \begin{tabular}{cccccccccc}
  \toprule
  problem & LAEA-Reg-8B & LAEA-8B & LAEA-8x7B & LAEA-Reg-8x7B & BO & SADE & SACC-EAM-II & SACOSO & SAMSO \\
  \midrule
  \multicolumn{10}{c}{$n=5$}  \\ \midrule
  \multirow{2}{*}{Ellipsoid} & 1.01e+00[3] & 2.30e+00[6]($\approx$) & 1.47e+00[5]($\approx$) & 2.52e+00[7]($-$) & \hl{6.15e-03[1]($+$)} & 5.72e+00[9]($-$) & 3.44e+00[8]($-$) & 1.20e+00[4]($\approx$) & 1.01e-01[2]($+$) \\ 
    & (9.29e-01) & (1.90e+00) & (1.26e+00) & (1.70e+00) & (2.65e-03) & (2.33e+00) & (1.86e+00) & (1.20e+00) & (1.32e-01) \\  \hline
  \multirow{2}{*}{Rosenbrock} & 8.35e+00[2] & 1.50e+01[5]($\approx$) & 1.38e+01[4]($-$) & 1.53e+01[6]($\approx$) & 1.08e+01[3]($\approx$) & 2.95e+01[9]($-$) & 1.65e+01[7]($-$) & 1.97e+01[8]($-$) & \hl{7.05e+00[1]($\approx$)} \\ 
    & (6.76e+00) & (1.53e+01) & (6.27e+00) & (9.22e+00) & (6.82e+00) & (1.53e+01) & (7.93e+00) & (5.19e+00) & (4.19e+00) \\  \hline
  \multirow{2}{*}{Ackley} & 6.83e+00[3] & 6.28e+00[2]($\approx$) & 7.76e+00[4]($\approx$) & 9.72e+00[6]($-$) & 1.76e+01[9]($-$) & 9.54e+00[5]($-$) & 1.21e+01[7]($-$) & \hl{5.18e+00[1]($\approx$)} & 1.31e+01[8]($-$) \\ 
    & (2.50e+00) & (2.35e+00) & (3.35e+00) & (2.89e+00) & (4.26e+00) & (1.69e+00) & (1.14e+00) & (1.14e+00) & (2.76e+00) \\  \hline
  \multirow{2}{*}{Griewank} & 5.10e+00[6] & 4.47e+00[5]($\approx$) & 2.74e+00[4]($\approx$) & 5.43e+00[7]($\approx$) & 2.28e+00[3]($\approx$) & 6.35e+00[9]($\approx$) & 6.08e+00[8]($\approx$) & 1.64e+00[2]($+$) & \hl{1.53e+00[1]($+$)} \\ 
    & (5.16e+00) & (5.05e+00) & (1.65e+00) & (3.28e+00) & (9.96e-01) & (3.77e+00) & (1.59e+00) & (1.63e+00) & (1.27e+00) \\  \midrule
    \multicolumn{10}{c}{$n=10$}  \\ \midrule
    \multirow{2}{*}{Ellipsoid} & 2.44e+01[4] & 2.51e+01[5]($\approx$) & 2.54e+01[6]($\approx$) & 3.90e+01[8]($\approx$) & \hl{9.29e-01[1]($+$)} & 7.22e+01[9]($-$) & 2.57e+01[7]($\approx$) & 1.60e+01[3]($+$) & 4.67e+00[2]($+$) \\ 
      & (7.98e+00) & (6.96e+00) & (1.76e+01) & (2.05e+01) & (2.61e+00) & (1.48e+01) & (1.54e+00) & (8.39e+00) & (4.65e+00) \\  \hline
    \multirow{2}{*}{Rosenbrock} & 9.41e+01[3] & 9.82e+01[4]($\approx$) & 1.70e+02[6]($\approx$) & 1.76e+02[7]($-$) & 1.23e+02[5]($\approx$) & 2.73e+02[9]($-$) & 1.88e+02[8]($-$) & 7.55e+01[2]($\approx$) & \hl{3.76e+01[1]($+$)} \\ 
      & (3.16e+01) & (2.77e+01) & (1.20e+02) & (9.54e+01) & (4.51e+01) & (9.73e+01) & (4.49e+01) & (5.49e+01) & (2.02e+01) \\  \hline
    \multirow{2}{*}{Ackley} & 1.18e+01[3] & 1.14e+01[2]($\approx$) & 1.37e+01[5]($\approx$) & 1.54e+01[6]($-$) & 1.73e+01[8]($-$) & 1.34e+01[4]($\approx$) & 1.66e+01[7]($-$) & \hl{5.87e+00[1]($+$)} & 1.76e+01[9]($-$) \\ 
      & (1.71e+00) & (1.45e+00) & (2.48e+00) & (2.54e+00) & (4.25e+00) & (1.25e+00) & (7.48e-01) & (1.50e+00) & (1.66e+00) \\  \hline
    \multirow{2}{*}{Griewank} & 1.44e+01[3] & 1.50e+01[4]($\approx$) & 2.48e+01[6]($\approx$) & 3.53e+01[7]($-$) & \hl{1.42e+00[1]($+$)} & 4.52e+01[9]($-$) & 3.94e+01[8]($-$) & 4.48e+00[2]($+$) & 2.06e+01[5]($\approx$) \\ 
      & (8.35e+00) & (5.91e+00) & (1.49e+01) & (1.61e+01) & (1.45e-01) & (1.38e+01) & (1.15e+01) & (3.01e+00) & (1.76e+01) \\  \hline
    mean rank& 3.375	&4.125&	5	&6.75	&3.875	&7.875&	7.5	&2.875	&3.625 \\
  $+$ / $-$ / $\approx$ &  & 0/0/8 & 0/1/7 & 0/5/3 & 3/2/3 & 0/6/2 & 0/6/2 & 4/1/3 & 4/2/2 \\ 
  \bottomrule
  \end{tabular}
  \label{tab:comparative}
  \end{table*}
% \end{sidewaystable*}

The results indicate that among the four implementations of LAEA, both LAEA-Reg-8B and LAEA-8B exhibit outstanding performance on most test problems, with no significant differences between them. In contrast, LAEA-8x7B and LAEA-Reg-8x7B perform relatively poorly. Considering that Llama3-8B* demonstrates better performance on regression tasks than on classification tasks, LAEA-Reg-8B, which uses Llama3-8B* solely as a regression model, has a superior average rank compared to LAEA-8B, which employs Llama3-8B* for both classification and regression tasks. Using LAEA-8B as a benchmark, in terms of average rank, LAEA-Reg-8B surpasses BO, SADE, SACC-EAM-II, and SAMSO algorithms, but performs worse than the SACOSO algorithm. According to the results of the Wilcoxon rank-sum test, LAEA-8B and BO exhibit respective advantages across the eight test problems, outperforming SADE and SACC-EAM-II, but falling short compared to SACOSO and SAMSO.

In summary, under a limited evaluation budget, LAEA achieves performance levels comparable to mainstream optimization algorithms when Llama3-8B* model is used solely for context-based inference.

\subsection{Model Performance in Pre-selection}

In section~\ref{sec:exp:comparative}, we utilized Llama3* and Mixtral-8x7B as surrogate models within the LAEA algorithm for comparisons with mainstream algorithms. To further evaluate the performance gains provided by LLMs and to eliminate the influence of additional model management strategies (e.g., solutions without evaluations as referenced in Algorithm~\ref{alg:llm_ea}), this section employs the CoDE algorithm~\cite{wang2011differential} as the basic framework. Figure~\ref{fig:preselection} illustrates the pre-selection process. CoDE is a multi-operator differential evolution algorithm where each parent generates multiple trial solutions. By using LLM to pre-select promising trial solutions, we can clearly evaluate the performance gains of using LLM as a surrogate model.
\begin{figure}[htbp]
\centering
\includegraphics[width=.7\linewidth]{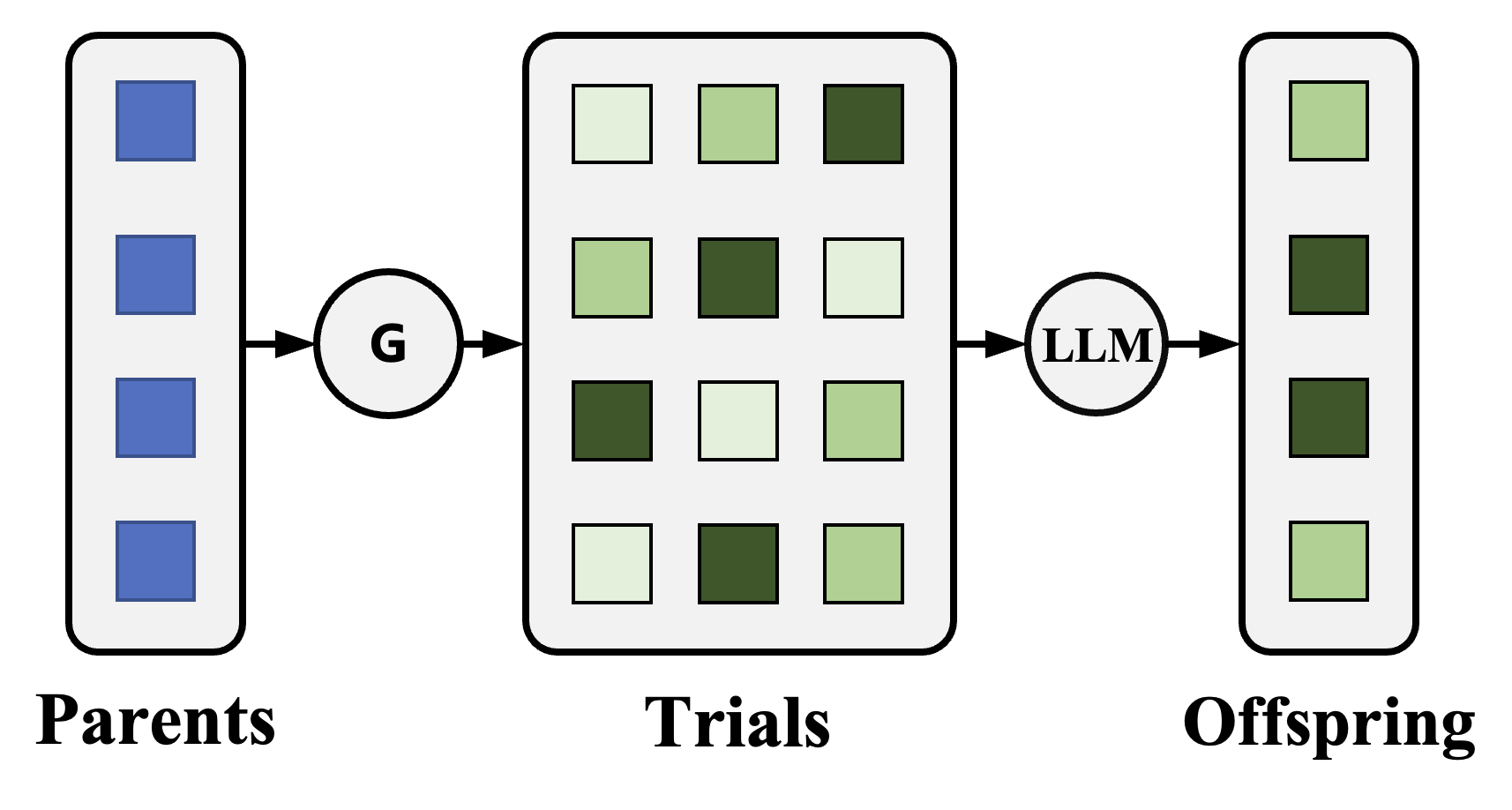}
\caption{Illustration of the pre-selection process in the CoDE algorithm.}
\label{fig:preselection}
\end{figure}
Specifically, each parent generates three trial solutions, and the LLM infers the quality of the solutions based on the parent population information. When using LLM as a regression model, the trials with the smallest predicted value is selected. When using LLM as a classification model, the offspring with the label ``1'' is selected (if multiple solutions exist, one is chosen randomly). Random selection is also used as a baseline algorithm to evaluate the performance of LLM.

Table~\ref{tab:preselection} presents the performance of Llama3* and Mixtral-8x7B in pre-selection tasks. The results show that, compared to random selection, both Llama3* and Mixtral-8x7B perform as well as or better than the baseline in pre-selection tasks. Specifically, when used as regression models, the two LLMs achieved better pre-selection performance, with average ranks of 1.25 and 1.75, respectively. When used as classification models, Llama3* and Mixtral-8x7B had average ranks of 3.375 and 3.625, respectively. These results indicate that LLMs can effectively select promising solutions through context-based inference, with better performance noted when LLMs are used as regression models. This finding is consistent with the results from previous experiments.

\begin{table}[htbp]
  \renewcommand{\arraystretch}{1.1}
  \renewcommand{\tabcolsep}{4pt}
  \centering
  \caption{Llama3* and Mixtral-8x7B performance in pre-selection tasks.}  \label{tab:preselection}
  \tiny
  \begin{tabular}{cccccc}
  \toprule
  problem & random & Llama3*-reg & Llama3*-cla & Mixtral-8x7B-reg & Mixtral-8x7B-cla \\
  \midrule
  \multicolumn{6}{c}{$n=5$}  \\ 
  \midrule
  \multirow{2}{*}{Ellipsoid} & 2.05e-01[5] & 1.87e-02[2]($+$) & 8.29e-02[3]($+$) & \hl{1.61e-02[1]($+$)} & 8.86e-02[4]($+$) \\ 
    & (1.13e-01) & (9.08e-03) & (4.20e-02) & (1.04e-02) & (8.61e-02) \\  \hline
  \multirow{2}{*}{Rosenbrock} & 5.12e+00[5] & \hl{2.22e+00[1]($+$)} & 4.42e+00[4]($\approx$) & 2.67e+00[2]($+$) & 3.99e+00[3]($+$) \\ 
    & (1.46e+00) & (8.59e-01) & (7.86e-01) & (9.38e-01) & (8.48e-01) \\  \hline
  \multirow{2}{*}{Ackley} & 4.34e+00[5] & \hl{2.53e+00[1]($+$)} & 3.68e+00[3]($+$) & 2.55e+00[2]($+$) & 3.89e+00[4]($\approx$) \\ 
    & (4.59e-01) & (5.36e-01) & (5.14e-01) & (5.18e-01) & (6.23e-01) \\  \hline
  \multirow{2}{*}{Griewank} & 1.14e+00[5] & 6.88e-01[2]($+$) & 9.58e-01[4]($\approx$) & \hl{6.32e-01[1]($+$)} & 8.85e-01[3]($+$) \\ 
    & (1.62e-01) & (1.60e-01) & (2.18e-01) & (1.28e-01) & (1.67e-01) \\    \midrule
    \multicolumn{6}{c}{$n=10$}  \\ 
    \midrule
    \multirow{2}{*}{Ellipsoid} & 1.80e+01[5] & \hl{6.41e+00[1]($+$)} & 1.57e+01[4]($\approx$) & 1.14e+01[2]($+$) & 1.25e+01[3]($\approx$) \\ 
      & (6.38e+00) & (2.41e+00) & (4.68e+00) & (3.97e+00) & (2.98e+00) \\  \hline
    \multirow{2}{*}{Rosenbrock} & 9.56e+01[5] & \hl{4.27e+01[1]($+$)} & 7.26e+01[3]($\approx$) & 6.01e+01[2]($+$) & 7.49e+01[4]($\approx$) \\ 
      & (2.56e+01) & (9.17e+00) & (2.52e+01) & (2.82e+01) & (1.48e+01) \\  \hline
    \multirow{2}{*}{Ackley} & 1.29e+01[5] & \hl{9.36e+00[1]($+$)} & 1.24e+01[3]($\approx$) & 1.07e+01[2]($+$) & 1.27e+01[4]($\approx$) \\ 
      & (6.73e-01) & (1.16e+00) & (9.02e-01) & (1.26e+00) & (9.34e-01) \\  \hline
    \multirow{2}{*}{Griewank} & 1.46e+01[5] & \hl{5.85e+00[1]($+$)} & 1.10e+01[3]($+$) & 8.29e+00[2]($+$) & 1.20e+01[4]($\approx$) \\ 
      & (3.23e+00) & (1.67e+00) & (2.74e+00) & (2.19e+00) & (3.40e+00) \\  \hline
  mean rank & 5.00 & 1.25 & 3.375 & 1.75 & 3.625 \\ 
  $+$ / $-$ / $\approx$ & & 8/0/0 & 3/0/5 & 8/0/0 & 3/0/5 \\ 
  \bottomrule
  \end{tabular}
  \end{table}

\subsection{Time Analysis}

This section analyzes the inference time overhead of LLMs under different scales of data. We selected the unquantized Llama3-8B* model and conducted inference time statistics on 5-dimensional and 10-dimensional data. The training dataset size is 50, and the testing dataset size is also 50. We recorded token counts and inference times for regression and classification prompts under different data precisions ($\beta$). The vllm platform was used as the backend for inference, with hardware comprising two NVIDIA 4090 GPUs. Both serial and parallel inference times were tested, with concurrency set to match the test dataset size.
The specific data are shown in Figure~\ref{fig:preselection}.

\begin{figure}[htbp]
  \centering
  \includegraphics[width=\linewidth]{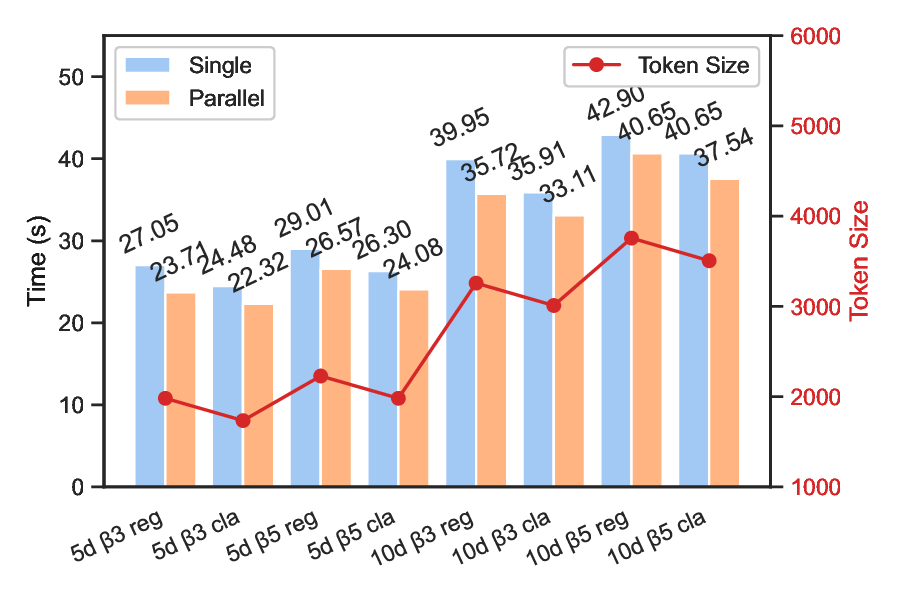}
  \caption{Inference time statistics of Llama3-8B* on different scales of data.}
  \label{fig:preselection}
\end{figure}

From the data, it can be observed that for 5-dimensional data, prompt lengths range between 3000-4000 tokens, with an inference time of approximately 20-30 seconds for 50 test samples. For 10-dimensional data, prompt lengths range between 4000-5000 tokens, with an inference time of approximately 30-40 seconds. Parallel inference times are reduced compared to serial ones but not linearly proportional to concurrency levels. Classification task prompts have fewer tokens than regression tasks; hence their corresponding inference times are relatively shorter. When $\beta=5$, token lengths are longer compared to when $\beta=3$, resulting in slightly increased inference times; however, this difference is marginal.

\section{Conclusion}
\label{sec:conclusion}

This work proposes a surrogate model that relies solely on inference from large language models (LLMs), aiming to assist selection within evolutionary algorithms and reduce dependence on the black-box function during the optimization process. Based on the paradigms of surrogate models in SAEAs, we define specific tasks for LLMs. By leveraging prompt engineering, we transform model-assisted selection into an inference task where LLMs assess the quality of candidate solutions based on historical evaluation data. Specifically, we implement classification and regression tasks to predict the category or approximate value of candidate solutions, thereby evaluating their quality. LLMs are then integrated as surrogate models within the evolutionary algorithm framework, resulting in the LLM-assisted Evolutionary Algorithm (LAEA). Initially, we evaluate the data selection capabilities of nine mainstream LLMs using low-dimensional visualization and fixed datasets. Subsequently, we conduct experiments on 5- and 10- dimensional test functions to test the performance of LAEA. The results indicate that LAEA is comparable to mainstream optimization algorithms under limited evaluation budgets. Additionally, we evaluate the performance of LLMs in pre-selection tasks, demonstrating that LLMs can effectively identify promising solutions through contextual inference. Finally, we analyze the inference time overhead of LLMs, showing that the overhead is acceptable across different data scales. Overall, this work demonstrates the feasibility of using LLMs as surrogate models in evolutionary algorithms, offering a novel approach to surrogate model construction. The code is available as an open-source repository for researchers.

However, some limitations are worth noting. For continuous optimization, traditional machine learning models remain the mainstream surrogate models due to several reasons. Firstly, the tokenizer used during LLM training may not handle numerical data well. Additionally, literature points to the limitations of self-attention and positional embedding in transformers, potentially resulting in insensitivity to numerical data~\cite{mcleish2024transformers}, which limits LLMs' efficacy in purely numerical problems. Secondly, the inference cost of LLMs cannot be overlooked. While using LLMs as surrogate models can avoid the training process of traditional surrogate models, the inference cost is significant, especially compared to efficient surrogate models like K-Nearest Neighbors (KNN) and random forests etc, where LLMs do not have an advantage in terms of time.

Despite these limitations, we remain optimistic about the broad application prospects of using LLMs as surrogate models in EAs. Future work can explore the following directions: 
\begin{itemize}
     \item Inference for Discrete or Non-numerical Data: In some industrial optimizations or human-computer interaction optimizations, the input and output of black-box models are not simple continuous values, posing significant challenges for traditional surrogate models. LLMs can handle these scenarios uniformly through embeddings. 
     \item Adapting LLMs for Numerical Contexts: Fine-tuning LLMs to be more adept in numerical scenarios can be achieved by modifying tokenizers or incorporating numerical data during LLM training, thereby enhancing their sensitivity to numerical data. 
     \item Optimizing Inference Time: Advanced inference frameworks have significantly improved the efficiency of LLMs. Furthermore, parallel inference for individual predictions can enhance prediction speed. Utilizing advanced computation and inference services like Groq\footnote{https://wow.groq.com/why-groq/} can also improve prediction speed. 
\end{itemize}

In conclusion, the application of LLMs as surrogate models in evolutionary algorithms holds promising prospects, although there are areas that need further refinement and optimization.

% \appendix
% \section{Example Appendix Section}


%%       \cite{<label>} ==> [1]

%%

\bibliographystyle{elsarticle-num} 
\bibliography{bare_jrnl.bib}




% %% For numbered reference style

%   \textit{\LaTeX: a document preparation system},
%   Addison Wesley, Massachusetts,

% \end{thebibliography}
\end{document}